\documentclass[10pt]{article}
\usepackage[letterpaper]{geometry}
\usepackage{hicss}
\usepackage{times}
\usepackage[none]{hyphenat}
\usepackage{url}
\usepackage{latexsym}
\usepackage{minted}
\usepackage{indentfirst}
\usepackage{graphicx}
\graphicspath{{images/}}
\usepackage[
    style=apa,
  ]{biblatex}
\title{Optimizing Class Distributions for Bias-Aware Multi-Class Learning}

\author{Mirco Felske \\
 CLAAS E-Systems GmbH \\
 {\underline{mirco.felske@claas.com}} \\ \\ \And
 Stefan Stiene \\
 Hochschule Osnabrück, University of Applied Sciences \\
 {\underline{s.stiene@hs-osnabrueck.de} } \\ \\}
\usepackage{multirow}
\usepackage{colortbl}
\usepackage[table]{xcolor}
\usepackage{algorithm}
\usepackage{algpseudocode}
\usepackage{amsmath}
\usepackage{cleveref}
\algtext*{EndFor}

\date{}

\begin{document}
\maketitle
\begin{abstract}
We propose BiCDO (Bias-Controlled Class Distribution Optimizer), an iterative, data-centric framework that identifies Pareto optimized class distributions for multi-class image classification. BiCDO enables performance prioritization for specific classes, which is useful in safety-critical scenarios (e.g. prioritizing 'Human' over 'Dog'). Unlike uniform distributions, BiCDO determines the optimal number of images per class to enhance reliability and minimize bias and variance in the objective function. BiCDO can be incorporated into existing training pipelines with minimal code changes and supports any labelled multi-class dataset. We have validated BiCDO using EfficientNet, ResNet and ConvNeXt on CIFAR-10 and iNaturalist21 datasets, demonstrating improved, balanced model performance through optimized data distribution.
\end{abstract}

\subsubsection*{Keywords:}

Biased-controlled dataset, Pareto optimized class-based dataset distribution

\section{Introduction}
\label{sec:introduction}
In the field of AI development, the focus is increasingly shifting from model-centric to data-centric artificial intelligence (\cite{Zha.2023, Salehi.2024, Jakubik.2024, Park.2024}). Like the model-centric approach, the data-centric approach tries to spread the data as evenly as possible among all classes. While new architectures are frequently developed to improve tasks such as classification, semantic segmentation and object detection, the dataset is often assumed to be equally distributed across the classes under consideration (\cite{Fan.2024, Hou.2023, Nah.2023}). Superficially, this is a valid assumption if all categories are more or less equally important (\cite{DuYingxiao.2023}). In an academic environment, this is often not questioned, as most datasets are already available with a balanced class distribution and thus conform to the prevailing assumption. In many industrial applications, the imbalance of the data is the first problem to be solved, since many real world data occur in a long-tail distribution (\cite{DuYingxiao.2023, Fan.2024, Hou.2023, XiaohuaChen.2023}). Previous approaches attempted to overcome this by re-weighting or re-sampling the data (\cite{Zhang.2024}). In our opinion, the reweighting approach is currently the best way to obtain an unbiased model because, unlike resampling, there is no potential for data corruption as in the oversampling approach and no omission of data as in the undersampling approach. However, even this approach does not lead to a completely unbiased model, as the weighting does not lead to sufficient variance in the training between the classes.
\begin{figure}[thb]
	\centering
	\includegraphics[width=\linewidth]{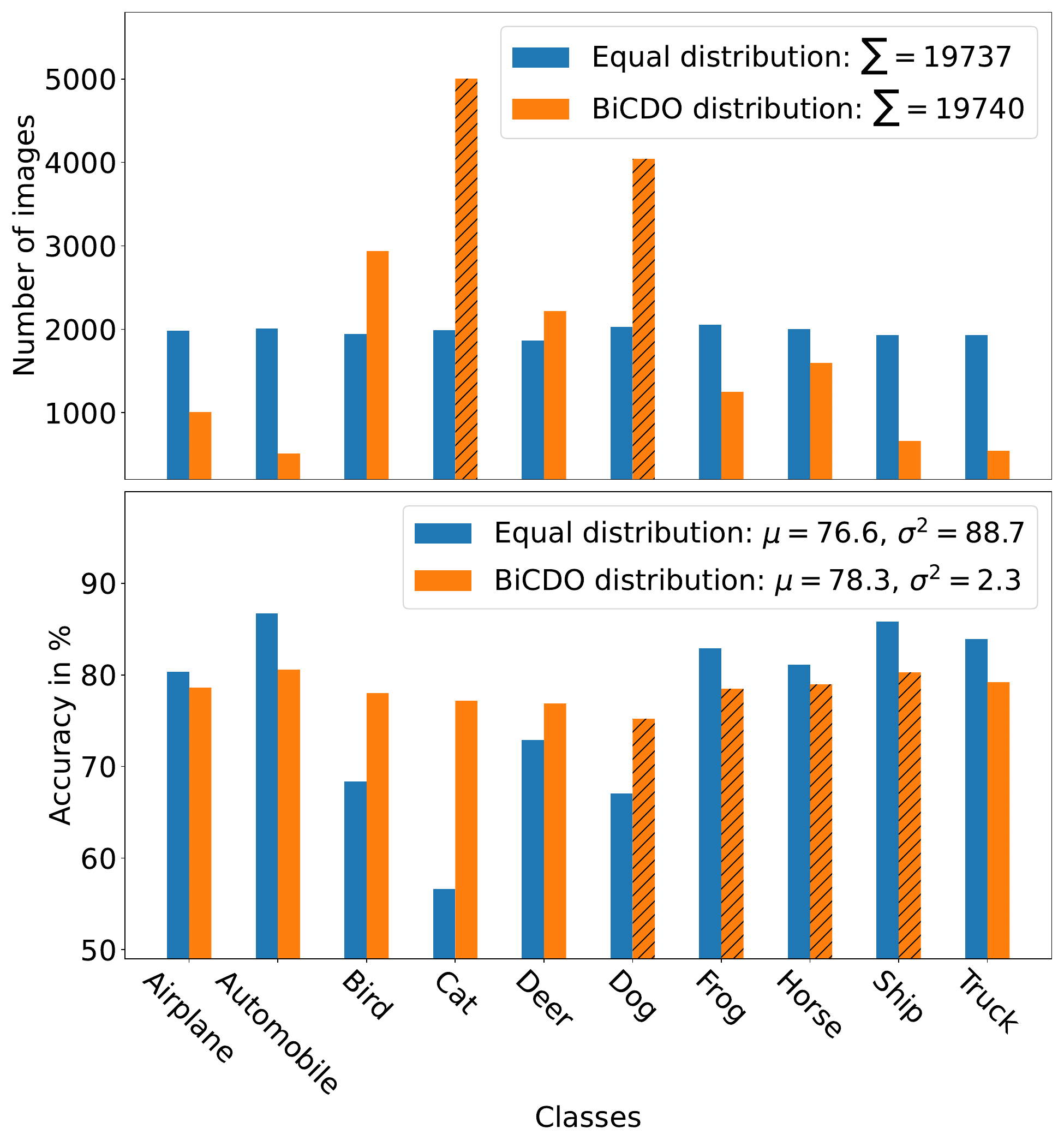}
	\caption{The upper image compares the class distributions of uniform and BiCDO-optimized CIFAR-10 with a similar total number of images. The lower image shows the accuracy, mean ($\boldsymbol{\mu}$) and variance ($\boldsymbol{\sigma^2}$) of EfficientNet-B3 model after training.} 
    \label{fig:introduction}
\end{figure}

\Cref{fig:introduction} illustrates that the current assumptions of a uniformly distributed dataset can lead to an imbalance in the evaluation. In particular, the variance of the accuracy for the uniformly distributed dataset is $88.7$. The BiCDO optimized dataset has a variance of $2.3$, which shows that it is possible to find the Pareto optimized number of images for each class in the dataset by aligning the objective function for each class to a certain level. In this paper, all tests are shown using the accuracy metric only, although other objective functions are possible. In all our experiments, a model trained on an BiCDO distributed dataset achieves comparable overall accuracy to a model trained on a uniformly distributed dataset. As can be seen in  \cref{fig:introduction}, the most notable difference is that the highest and lowest accuracy values of the model trained with the BiCDO distributed data are almost the same. In contrast, the variance of accuracy is more than 30 times higher for the model trained on uniformly distributed data. This demonstrates that having the same number of images for all classes is not a prerequisite for developing optimal and reliable models. Therefore, it is necessary to identify the combination of the Pareto optimized data used and the model to be trained. Using this approach, it is found that the accuracy values for all classes are almost identical, providing evidence that the model is not biased towards any particular class. In this way, BiCDO shows the absence of overfitting to a particular class, which is an important objective for example in safety-critical applications. In addition, BiCDO provides insight into which class may require more data to avoid underfitting. This enables a highly efficient, objective-oriented data collection process, as the amount and type of data required to train a representative and unbiased model can be accurately predicted before further data is collected.

In \cref{subsec:find_ratios}, experiments show that BiCDO can be used to find the optimized class-based data distribution within a multi-class dataset. \Cref{subsec:expand_ratios} shows that a class-based offset can be used to obtain class-specific performance improvements. The framework is independent of the objective function. It can be used with any objective function as long as it can be described by a single value and does not have conflicting objectives. For example, the F1 score with internal calculation of precision and recall would not yield meaningful distribution factors.\\
In summary, this work provides the following contributions:
\begin{itemize}
    \item A novel iterative approach, called Bias-Controlled Class Distribution Optimizer, which is able to identify the Pareto optimized class-based data distribution of a given dataset and model.  
    \item The proof that a balanced dataset does not necessarily lead to a balanced result in terms of the objective function, but that this goal requires a systematic long-tail distribution of the classes in the dataset.
    \item Experiments showing that the found distribution can be equally extended to optimize the performance of a model.
\end{itemize}
In this paper, the term "Pareto-optimized" is used informally to clarify the objective of BiCDO. Since no proof of multi-objective optimality is provided, BiCDO should be regarded as a variance-equalizing heuristic that, under favorable conditions, can yield a Pareto-like class distribution. Convergence to this solution requires sufficient images for each class, limits that are not simultaneously reached, and an underlying model capable of converging on the data.
\begin{figure*}[tbh]
  \centering
  \includegraphics[width=\textwidth]{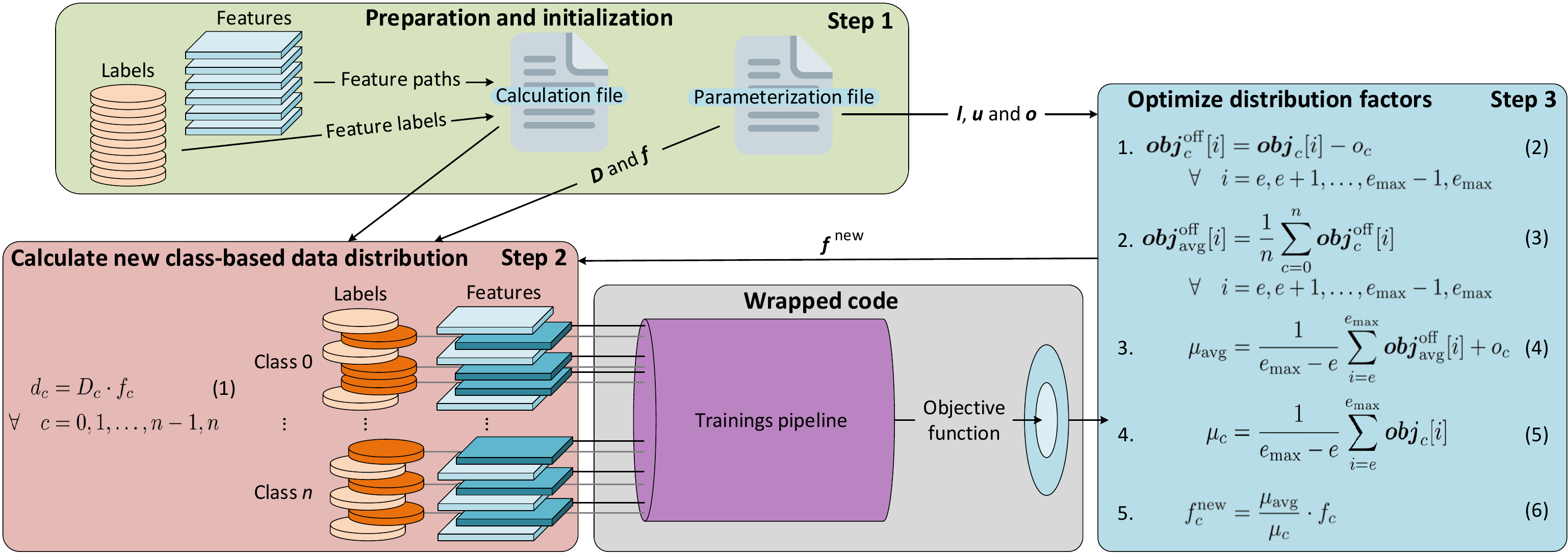}
    \caption{The figure shows the BiCDO workflow and how it wraps around an existing training pipeline.}
    \label{fig:workflow_optms}
\end{figure*}
\section{Related Work}
\label{sec:related_work}
\vspace{-0.2em}
Previous work has considered the problem of long-tailed datasets. As mentioned above, in most cases the problem is attempted to be solved by re-sampling or re-weighting (\cite{Zhang.2024}). Re-sampling aims to either undersample the head classes (\cite{Tahir.2012,Zang.2021}) or oversample (\cite{Singh.2018,Yan.2019,Park.2022}) the tail classes. Re-weighting uses the loss function to give more weight to certain classes (\cite{Cao.2019,Tan.2020,Hong.2021,Samuel.2021,Li.2022}). There are other methods that can be derived from this, such as augmentation (\cite{Li.2021,He.2021,Liu.2021,Wang.2021,Chen.2022}), decoupled training (\cite{Zhang.2021,Zhong.2021,Desai.2021}) and ensemble learning (\cite{Wang.2020,Guo.2021,Zhang.2022}).
The objective of these approaches is to address the limitations of long-tail datasets by modifying the training of the model or the data itself in order to ensure that the model is trained in an unbiased manner.

If these approaches are to be used, it must first be determined to what extent they actually achieve the goal of training an unbiased model. As \cite{Sinha.2020} and \cite{Sinha.2022} show for long-tail data, there is not always a correlation between the number of samples and class-specific performance. According to \cite{Ma.2022}, this correlation does not exist for non-long-tailed data either. Also \cite{Ma.2023} builds on the result of \cite{Ma.2022} that there is not always a correlation between the number of samples and class-specific performance by analysing the fairness of the models from a geometric perspective. The assumption that more uniformly distributed data does not lead to a better result is also used by \cite{JayGala.2024}. They use a generative adversarial network (\cite{Goodfellow.10.06.2014}) to generate synthetic data to provide additional examples for the classes that are actually difficult for this classifier to learn. We agree that more uniformly distributed data does not necessarily lead to a better result. 

However, our method does not aim to optimize multiple objective functions, as in MGDA \cite{Desideri.2012} or EPO \cite{Mahapatra.2021}, nor does it aim to minimize the variance of the underlying non-normally distributed data, as in auto-lambda \cite{LiuShikun.2022}. It is also not an adaptive sampling method that tries to distribute the weights for the data in such a way that they are efficient without loss of information and with as little data as possible, as in the learn2mix algorithm from \cite{Venkatasubramanian.2024}. Furthermore, the goal is not to quantify which class is learned worse or better, as in \cite{Ma.2023}, nor to obtain additional training data via an automated, possibly biased network, as in \cite{JayGala.2024}. Rather, we want to show that BiCDO further strengthens the data-centric perspective by attempting to map the variance of an object over the number of objects in the dataset in order to minimize the variance of the objective function between objects.
\section{Method}
\label{sec:method}
The overall approach is based on the assumption that the result of the objective function of a single class will be better if quantitatively more data per class is provided during training, and vice versa. The amount of data per class is adjusted by class-specific scalar hyperparameters $f_c$. To avoid a time-consuming hyperparameter search, the gradient of $f_c$ is directly approximated by the first part of \cref{eq:fact_new}, which is the reciprocal normalized result of the class-specific objective function. Multiplying the hyperparameters $f_c$ from the previous iteration by the approximated gradient results in a reduction of the inter-class variance in the next iteration. The hyperparameter optimization converges until no single $f_c$ can be optimized further by sacrificing a higher inter-class variance, and thus approaches a Pareto optimized solution, or reaches the BiCDO limits $l_c$ and $u_c$. If it converges, any change in the amount of data for one class will degrade the accuracy of the other classes (cf. Tab. 1 (B).2k and Tab. 3 (A)). Hence the name Bias-Controlled Class Distribution Optimizer (BiCDO). As shown in \cref{subsec:find_ratios}, the purpose of the distribution in the first step is to minimize the variance between classes in the objective function, and thus to show that it leads to a bias-controlled, fairly trained model. Subsequently, \cref{subsec:expand_ratios} shows that the Pareto optimized class-based data distribution found can be extended to improve the overall performance. As can be seen in \cref{fig:workflow_optms}, BiCDO consists of three steps. For easier understanding and implementation, the algorithm is presented in algorithm 1 as pseudo-code.
\begin{algorithm}
\caption{Pseudo-code of the BiCDO algorithm}
\label{alg:bicdo}
\begin{algorithmic}[1]
\State \textbf{Preparation and initialization}. 
       \Statex \hspace*{\algorithmicindent}Define class-specific parameters: 
       \Statex \hspace*{\algorithmicindent}$D_c, f_c, l_c, u_c, o_c$.
\Statex \textbf{for} each iteration (e.g., 150 steps) \textbf{do}
    \State \hspace*{\algorithmicindent}\textbf{Data provision and training:} 
        \Statex \hspace*{\algorithmicindent}\hspace*{\algorithmicindent}Select samples per class using \Cref{eq:it_quant}.
        \Statex \hspace*{\algorithmicindent}\hspace*{\algorithmicindent}Train the model and compute class \hspace*{\algorithmicindent}\hspace*{\algorithmicindent}performance.
    \State \hspace*{\algorithmicindent}\textbf{Distribution optimization:}
        \Statex \hspace*{\algorithmicindent}\hspace*{\algorithmicindent}Update class factors using \Cref{eq:fact_new}.
        \Statex \hspace*{\algorithmicindent}\hspace*{\algorithmicindent}Set $f_c \gets f_c^{\text{new}}$ for next iteration.
\end{algorithmic}
\end{algorithm}

\textbf{Step one} is data preparation and initialization. In this step, all available data is read in to store the path and label of each image in a calculation file. All further calculations are based on this file to find the optimized data distribution. Within the initialization of BiCDO, the parameters: maximum available data $D_c$, a distribution factor $f_c$ to define the amount of data for the first iteration, a lower and an upper limit ($l_c$ and $u_c$ between zero and one) for the number of images provided to the training algorithm and a target offset $o_c$ for the optimization function have to be defined for each class $c$ in the parameterization file. The target offset $o_c$ can be used to tell BiCDO that certain classes should have a better or worse objective function value relative to the average objective function value. The parameter must be between $-1$ and $1$. The initialization parameters are used for optimization in steps two and three.

\textbf{Step two} is to provide the data. With the number of classes $n$, the maximum available data $D_c$ and the factor $f_c$ between zero and one, the quantity
\begin{align}\label{eq:it_quant}
 d_c &= D_c \cdot f_c\\
 \forall\quad c&=0,1,\dots,n-1,n\nonumber
\end{align}
\begin{figure}[tbh]
	\centering
	\includegraphics[width=\linewidth]{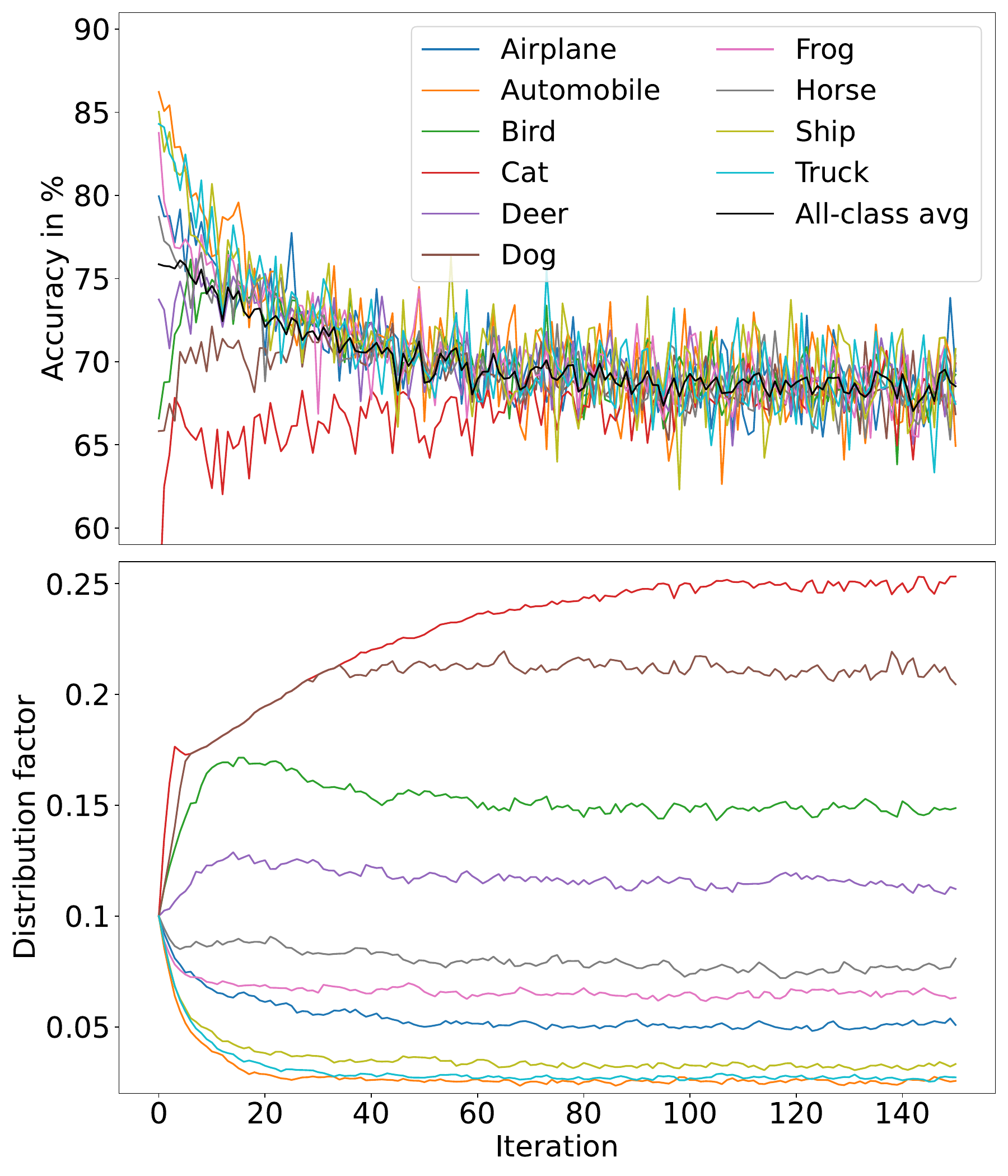}
	\caption{The upper image illustrates the changes in average accuracy ($\boldsymbol{\mu_{\mathrm{avg}}}$) and per-class accuracy ($\boldsymbol{\mu_c}$) across iterations. The lower image shows how the class distribution factors ($\boldsymbol{f_c}$) evolved.} 
    \label{fig:experiments_find}
\end{figure}
to be provided for this iteration per class $c$ is calculated. A seed is used to randomly select training images to represent the previously calculated amount of data for each class. The data is then fed into the actual training algorithm. The adaptation of the dataset is only applied to the training images. The set of validation images remains unchanged. The model is trained within the training algorithm using all the training data provided. The objective function $\boldsymbol{obj}_c$ (e.g. accuracy, precision or recall) is calculated on the validation dataset for each class and for each epoch. Starting from a defined epoch $e$ and ending with the last epoch $e_{max}$, so that only the epochs in which the objective function has converged are taken into account. This allows BiCDO to converge faster. The result is returned to BiCDO, which then calculates the optimal amount of data for each class for this iteration.

\begin{table*}[tbh]
	\caption{The table presents rounded CIFAR-10 results for varying image counts per class and model architectures. $\boldsymbol{D_c}$ is the max images per class, $\boldsymbol{f}$ the distribution factor, and $\boldsymbol{acc}$ the accuracy ($\boldsymbol{\%}$).}
    \centering
    \small
    \makebox[\linewidth]{
\begin{tabular}{l|ll||cccccccccccc}
\multicolumn{3}{c}{} & \multicolumn{3}{c}{(A)} & \multicolumn{3}{c}{(B)} & \multicolumn{3}{c}{(C)} & \multicolumn{3}{c}{(D)} \\
\multicolumn{3}{c||}{Model} & \multicolumn{3}{c|}{EfficientNet b0} & \multicolumn{3}{c|}{EfficientNet b3} & \multicolumn{3}{c|}{ResNet} & \multicolumn{3}{c}{ConNeXt}\\ \hline 
\multicolumn{3}{c||}{$D_c$} & \multicolumn{1}{c}{0.5k} & \multicolumn{1}{c}{2k} & \multicolumn{1}{c|}{5k} & \multicolumn{1}{c}{0.5k} & \multicolumn{1}{c}{\cellcolor{red!40}2k} & \multicolumn{1}{c|}{5k} & \multicolumn{1}{c}{0.5k} & \multicolumn{1}{c}{2k} & \multicolumn{1}{c|}{5k} & \multicolumn{1}{c}{0.5k} & \multicolumn{1}{c}{2k} & \multicolumn{1}{c}{5k} \\ \hline\hline
\multirow{20}{*}{\rotatebox[origin=c]{90}{Class results}} & \multirow{2}{*}{Airplane} & $acc$ & $58.4$ & $68.5$ & \multicolumn{1}{c|}{$76.3$} & $58.3$ & \cellcolor{red!40}$69.4$ & \multicolumn{1}{c|}{$69.6$} & $59.8$ & $72.2$ & \multicolumn{1}{c|}{$78.2$} & $58.4$ & $67.6$ & $80.6$\\
&& $f$ & $0.06$ & $0.05$ & \multicolumn{1}{c|}{$0.05$} & $0.06$ & \cellcolor{red!40}$0.05$ & \multicolumn{1}{c|}{$0.04$} & $0.06$ & $0.06$ & \multicolumn{1}{c|}{$0.05$} & $0.06$ & $0.06$ & $0.05$ \\ \cline{3-15}
& \multirow{2}{*}{Automobile} & $acc$ & $60$ & $72.8$ & \multicolumn{1}{c|}{$78.3$} & $56.3$ & \cellcolor{red!40}$64.9$ & \multicolumn{1}{c|}{$69.7$} & $63.8$ & $71.7$ & \multicolumn{1}{c|}{$79.7$} & $55.5$ & $63.1$ & $79.7$ \\
&& \cellcolor{blue!40}$f$ & \cellcolor{blue!40}$0.04$ & \cellcolor{blue!40}$0.02$ & \multicolumn{1}{c|}{\cellcolor{blue!40}$0.02$} & \cellcolor{blue!40}$0.04$ & \cellcolor{red!40}$0.03$ & \multicolumn{1}{c|}{\cellcolor{blue!40}$0.02$} & \cellcolor{blue!40}$0.04$ & \cellcolor{blue!40}$0.03$ & \multicolumn{1}{c|}{\cellcolor{blue!40}$0.02$} & \cellcolor{blue!40}$0.04$ & \cellcolor{blue!40}$0.02$ & \cellcolor{blue!40}$0.02$ \\ \cline{3-15}
& \multirow{2}{*}{Bird} & $acc$ & $62.2$ & $70$ & \multicolumn{1}{c|}{$75.8$} & $59.5$ & \cellcolor{red!40}$69.2$ & \multicolumn{1}{c|}{$67.9$} & $62.2$ & $74.4$ & \multicolumn{1}{c|}{$78.4$} & $56.5$ & $63.6$ & $81.6$ \\
&& $f$ & $0.17$ & $0.14$ & \multicolumn{1}{c|}{$0.14$} & $0.16$ & \cellcolor{red!40}$0.15$ & \multicolumn{1}{c|}{$0.15$} & $0.15$ & $0.14$ & \multicolumn{1}{c|}{$0.13$} & $0.15$ & $0.15$ & $0.13$ \\ \cline{3-15}
& \multirow{2}{*}{Cat} & $acc$ & $58.5$ & $69.8$ & \multicolumn{1}{c|}{$76.2$} & $53.4$ & \cellcolor{red!40}$69.9$ & \multicolumn{1}{c|}{$68.5$} & $62.8$ & $72.3$ & \multicolumn{1}{c|}{$76.8$} & $55.4$ & $68.5$ & $81$ \\
&& \cellcolor{blue!40}$f$ & \cellcolor{blue!40}$0.2$ & \cellcolor{blue!40}$0.25$ & \multicolumn{1}{c|}{\cellcolor{blue!40}$0.26$} & \cellcolor{blue!40}$0.2$ & \cellcolor{red!40}$0.25$ & \multicolumn{1}{c|}{\cellcolor{blue!40}$0.26$} & \cellcolor{blue!40}$0.22$ & \cellcolor{blue!40}$0.24$ & \multicolumn{1}{c|}{\cellcolor{blue!40}$0.26$} & \cellcolor{blue!40}$0.21$ & \cellcolor{blue!40}$0.25$ & \cellcolor{blue!40}$0.27$ \\ \cline{3-15}
& \multirow{2}{*}{Deer} & $acc$ & $56.7$ & $70.8$ & \multicolumn{1}{c|}{$76.4$} & $56.7$ & \cellcolor{red!40}$68.7$ & \multicolumn{1}{c|}{$70.2$} & $62.8$ & $73.9$ & \multicolumn{1}{c|}{$79$} & $60.6$ & $67.9$ & $81$ \\
&& $f$ & $0.12$ & $0.11$ & \multicolumn{1}{c|}{$0.11$} & $0.12$ & \cellcolor{red!40}$0.11$ & \multicolumn{1}{c|}{$0.12$} & $0.12$ & $0.11$ & \multicolumn{1}{c|}{$0.11$} & $0.13$ & $0.12$ & $0.1$ \\ \cline{3-15}
& \multirow{2}{*}{Dog} & $acc$ & $61.5$ & $70.5$ & \multicolumn{1}{c|}{$76$} & $57$ & \cellcolor{red!40}$66.8$ & \multicolumn{1}{c|}{$69.4$} & $63.4$ & $70.6$ & \multicolumn{1}{c|}{$79.3$} & $57.4$ & $65.6$ & $78.7$ \\
&& $f$ & $0.18$ & $0.21$ & \multicolumn{1}{c|}{$0.23$} & $0.18$ & \cellcolor{red!40}$0.2$ & \multicolumn{1}{c|}{$0.22$} & $0.19$ & $0.21$ & \multicolumn{1}{c|}{$0.25$} & $0.18$ & $0.19$ & $0.24$ \\ \cline{3-15}
& \multirow{2}{*}{Frog} & $acc$ & $57.2$ & $69.5$ & \multicolumn{1}{c|}{$74.4$} & $59.3$ & \cellcolor{red!40}$67.6$ & \multicolumn{1}{c|}{$68.4$} & $65.4$ & $73.8$ & \multicolumn{1}{c|}{$77$} & $56.6$ & $69.6$ & $80.7$ \\
&& $f$ & $0.07$ & $0.06$ & \multicolumn{1}{c|}{$0.06$} & $0.08$ & \cellcolor{red!40}$0.06$ & \multicolumn{1}{c|}{$0.07$} & $0.07$ & $0.06$ & \multicolumn{1}{c|}{$0.06$} & $0.07$ & $0.06$ & $0.06$ \\ \cline{3-15}
& \multirow{2}{*}{Horse} & $acc$ & $58.7$ & $73$ & \multicolumn{1}{c|}{$76.5$} & $54.9$ & \cellcolor{red!40}$70.4$ & \multicolumn{1}{c|}{$68.5$} & $63.5$ & $73.6$ & \multicolumn{1}{c|}{$78.3$} & $60.8$ & $66.2$ & $77.8$ \\
&& $f$ & $0.08$ & $0.08$ & \multicolumn{1}{c|}{$0.08$} & $0.08$ & \cellcolor{red!40}$0.08$ & \multicolumn{1}{c|}{$0.07$} & $0.08$ & $0.09$ & \multicolumn{1}{c|}{$0.08$} & $0.08$ & $0.07$ & $0.07$ \\ \cline{3-15}
& \multirow{2}{*}{Ship} & $acc$ & $59.2$ & $71.9$ & \multicolumn{1}{c|}{$73.6$} & $57.5$ & \cellcolor{red!40}$70.8$ & \multicolumn{1}{c|}{$69.5$} & $67.6$ & $73.2$ & \multicolumn{1}{c|}{$79.1$} & $58.9$ & $68.1$ & $79.6$ \\
&& $f$ & $0.05$ & $0.03$ & \multicolumn{1}{c|}{$0.03$} & $0.05$ & \cellcolor{red!40}$0.03$ & \multicolumn{1}{c|}{$0.03$} & $0.04$ & $0.03$ & \multicolumn{1}{c|}{$0.03$} & $0.04$ & $0.03$ & $0.03$ \\ \cline{3-15}
& \multirow{2}{*}{Truck} & $acc$ & $63.5$ & $69.7$ & \multicolumn{1}{c|}{$77.9$} & $55.5$ & \cellcolor{red!40}$67.5$ & \multicolumn{1}{c|}{$70.4$} & $60.8$ & $71.7$ & \multicolumn{1}{c|}{$78.7$} & $55.2$ & $67.3$ & $80.4$ \\
&& $f$ & $0.04$ & $0.03$ & \multicolumn{1}{c|}{$0.02$} & $0.04$ & \cellcolor{red!40}$0.03$ & \multicolumn{1}{c|}{$0.02$} & $0.04$ & $0.03$ & \multicolumn{1}{c|}{$0.02$} & $0.04$ & $0.03$ & $0.02$ \\ \hline\hline 
\multicolumn{2}{l}{All-class average} & $\mu_{\mathrm{acc}}$ & $59.6$ & $70.7$ & \multicolumn{1}{c|}{$76.1$} & $56.8$ & \cellcolor{red!40}$68.5$ & \multicolumn{1}{c|}{$69.2$} & $63.2$ & $72.7$ & \multicolumn{1}{c|}{$78.5$} & $57.5$ & $66.6$ & \multicolumn{1}{c}{$80.1$}\\
\multicolumn{2}{l}{All-class variance} & $\sigma^2_{\mathrm{acc}}$ & $4.4$ & $2$ & \multicolumn{1}{c|}{$1.8$} & $3.4$ & \cellcolor{red!40}$3$ & \multicolumn{1}{c|}{$0.6$} & $4.3$ & $1.3$ & \multicolumn{1}{c|}{$0.8$} & $3.9$ & $3.9$ & \multicolumn{1}{c}{$1.2$}
\end{tabular}}
	\label{tab:find_ratios_cifar10}
\end{table*}
\textbf{Step three} consists of the optimization function, which is the basis for calculating the distribution factor $f_c^{\mathrm{new}}$ for the next iteration. The optimization function uses vectors to calculate the factor for each class and then determines the number of images to provide for each class. To get the offset $o_c$ added objective function
\begin{align}\label{eq:off_added_acc_c}
 \boldsymbol{obj}^{\mathrm{off}}_c[i]&=\boldsymbol{obj}_c[i]-o_c\\
 \forall\quad i&=e,e+1,\dots,e_{\mathrm{max}}-1,e_{\mathrm{max}}\nonumber
\end{align}
the target offset $o_c$ is subtracted from the objective of each class $\boldsymbol{obj}_c$ over all considered epochs from $e$ to $e_{max}$. This suggests that the objective function $\boldsymbol{obj}^{\mathrm{off}}_c$ for a particular class is worse or better than it actually is when the offset $o_c$ is given (not zero). In addition, the offset is taken into account in the following calculation of the average objective function
\begin{align}\label{eq:off_added_acc_a}
 \boldsymbol{obj}^{\mathrm{off}}_{\mathrm{avg}}[i]&=\frac{1}{n}\sum_{c=0}^n\boldsymbol{obj}^{\mathrm{off}}_c[i]\\
 \forall\quad i&=e,e+1,\dots,e_{\mathrm{max}}-1,e_{\mathrm{max}}\nonumber
\end{align}
across all classes. The offset-free average over all corresponding epochs 
\begin{align}\label{eq:mean_a}
 \mu_{\mathrm{avg}}=\frac{1}{e_{\mathrm{max}}-e}\sum_{i=e}^{e_{\mathrm{max}}}\boldsymbol{obj}^{\mathrm{off}}_{\mathrm{avg}}[i]+o_c
\end{align}
is then calculated by displaying it as a vector and adding the target offset $o_c$ so that the number of images is correctly adjusted. The class-specific averages
\begin{align}\label{eq:mean_c}
 \mu_c&=\frac{1}{e_{\mathrm{max}}-e}\sum_{i=e}^{e_{\mathrm{max}}}\boldsymbol{obj}_c[i]
\end{align}
are needed to divide the offset-free average $\mu_{\mathrm{avg}}$ by the class-specific averages $\mu_c$. Finally, the results are multiplied by the old factors $f_c$ to determine the factors
\begin{align}\label{eq:fact_new}
 f^{\mathrm{new}}_c&=\frac{\mu_{\mathrm{avg}}}{\mu_c}\cdot f_c
\end{align}
for the next iteration for one class. Doing this for all classes will result in an adjustment for all classes. All classes that were harder to learn in terms of the objective function will get a higher factor $f^{\mathrm{new}}_c$ and all classes that were supposedly easier to learn will get a lower factor $f^{\mathrm{new}}_c$ than in the previous iteration. In order not to exceed the limits of the maximum possible data volume, the resulting values are compared with the specified limits and adjusted if necessary. Once one class has reached its lower bound $l_c$ and another class has reached its upper bound $u_c$, there is not enough data to compute the Pareto optimized class-based data distribution. The maximum available data per class $D_c$ is then used to determine the number of images to provide for each class. The next iteration can then be started with step 2 and $f_c = f^{\mathrm{new}}_c$.

Steps two and three are repeated iteratively. The optimization function adjusts the training distribution after each training run so that the next, newly trained model has a more similar objective score across all classes after its training run than the previously trained model. Classes with a non-zero target offset are an exception. These classes will have the average objective score of all classes plus their target offset.

\subsection{Pareto Optimized Distribution}
\label{subsec:find_ratios}
\begin{table*}[tbh]
	\caption{The table shows rounded iNaturalist21 results for varying images per class and model architectures. $\boldsymbol{D_c}$ is the max images per class, $\boldsymbol{f}$ the distribution factor, and $\boldsymbol{acc}$ the accuracy ($\boldsymbol{\%}$).}
    \centering
    \small
    \makebox[\linewidth]{
\begin{tabular}{l|ll||cccccccc}
\multicolumn{3}{c}{} & \multicolumn{2}{c}{(A)} & \multicolumn{2}{c}{(B)} & \multicolumn{2}{c}{(D)} & \multicolumn{2}{c}{(E)} \\
\multicolumn{3}{c||}{Model} & \multicolumn{2}{c|}{EfficientNet b0} & \multicolumn{2}{c|}{EfficientNet b3} & \multicolumn{2}{c|}{ResNet} & \multicolumn{2}{c}{ConNeXt}\\ \hline 
\multicolumn{3}{c||}{$D_c$} & \multicolumn{1}{c}{0.5k} & \multicolumn{1}{c|}{2k} & \multicolumn{1}{c}{0.5k} & \multicolumn{1}{c|}{2k} & \multicolumn{1}{c}{0.5k} & \multicolumn{1}{c|}{2k} & \multicolumn{1}{c}{0.5k} & \multicolumn{1}{c}{2k} \\ \hline\hline
\multirow{20}{*}{\rotatebox[origin=c]{90}{Class results}} & \multirow{2}{*}{Plants} & $acc$ & $44.3$ & \multicolumn{1}{c|}{$49.2$} & $47$ & \multicolumn{1}{c|}{$50.7$} & $35.2$ & \multicolumn{1}{c|}{$43.9$} & $34.9$ & $38.9$ \\
&& \cellcolor{blue!40}$f$ & \cellcolor{blue!40}$0.03$ & \multicolumn{1}{c|}{\cellcolor{blue!40}$0.02$} & \cellcolor{blue!40}$0.03$ & \multicolumn{1}{c|}{\cellcolor{blue!40}$0.03$} & \cellcolor{blue!40}$0.03$ & \multicolumn{1}{c|}{\cellcolor{blue!40}$0.03$} & \cellcolor{blue!40}$0.02$ & \cellcolor{blue!40}$0.02$ \\ \cline{3-11}
& \multirow{2}{*}{Insects} & $acc$ & $41.6$ & \multicolumn{1}{c|}{$49.3$} & $41.3$ & \multicolumn{1}{c|}{$52.5$} & $38.4$ & \multicolumn{1}{c|}{$43.6$} & $30.9$ & $40$ \\
&& $f$ & $0.09$ & \multicolumn{1}{c|}{$0.09$} & $0.09$ & \multicolumn{1}{c|}{$0.09$} & $0.09$ & \multicolumn{1}{c|}{$0.08$} & $0.09$ & $0.08$ \\ \cline{3-11}
& \multirow{2}{*}{Birds} & $acc$ & $41.3$ & \multicolumn{1}{c|}{$50.7$} & $44.1$ & \multicolumn{1}{c|}{$50.8$} & $36.8$ & \multicolumn{1}{c|}{$44.7$} & $34$ & $39.4$ \\
&& \cellcolor{blue!40}$f$ & \cellcolor{blue!40}$0.05$ & \multicolumn{1}{c|}{\cellcolor{blue!40}$0.04$} & \cellcolor{blue!40}$0.05$ & \multicolumn{1}{c|}{\cellcolor{blue!40}$0.05$} & \cellcolor{blue!40}$0.05$ & \multicolumn{1}{c|}{\cellcolor{blue!40}$0.05$} & \cellcolor{blue!40}$0.06$ & \cellcolor{blue!40}$0.05$ \\ \cline{3-11}
& \multirow{2}{*}{Fungi} & $acc$ & $41.2$ & \multicolumn{1}{c|}{$53.9$} & $42.3$ & \multicolumn{1}{c|}{$57.3$} & $37$ & \multicolumn{1}{c|}{$47.4$} & $36.8$ & $38.7$ \\
&& \cellcolor{blue!40}$f$ & \cellcolor{blue!40}$0.05$ & \multicolumn{1}{c|}{\cellcolor{blue!40}$0.05$} & \cellcolor{blue!40}$0.05$ & \multicolumn{1}{c|}{\cellcolor{blue!40}$0.05$} & \cellcolor{blue!40}$0.05$ & \multicolumn{1}{c|}{\cellcolor{blue!40}$0.05$} & \cellcolor{blue!40}$0.06$ & \cellcolor{blue!40}$0.04$ \\ \cline{3-11}
& \multirow{2}{*}{Reptiles} & $acc$ & $42.8$ & \multicolumn{1}{c|}{$50.4$} & $47.6$ & \multicolumn{1}{c|}{$54$} & $38.3$ & \multicolumn{1}{c|}{$50.1$} & $35.1$ & $42.5$ \\
&& \cellcolor{red!40}$f$ & \cellcolor{red!40}$0.15$ & \multicolumn{1}{c|}{\cellcolor{red!40}$0.15$} & \cellcolor{red!40}$0.15$ & \multicolumn{1}{c|}{\cellcolor{red!40}$0.15$} & \cellcolor{red!40}$0.15$ & \multicolumn{1}{c|}{\cellcolor{red!40}$0.14$} & \cellcolor{red!40}$0.15$ & \cellcolor{red!40}$0.15$ \\ \cline{3-11}
& \multirow{2}{*}{Mammals} & $acc$ & $47.8$ & \multicolumn{1}{c|}{$52.2$} & $43.8$ & \multicolumn{1}{c|}{$53$} & $40.1$ & \multicolumn{1}{c|}{$48.6$} & $37.5$ & $38.3$ \\
&& $f$ & $0.07$ & \multicolumn{1}{c|}{$0.06$} & $0.07$ & \multicolumn{1}{c|}{$0.07$} & $0.07$ & \multicolumn{1}{c|}{$0.07$} & $0.08$ & $0.07$ \\ \cline{3-11}
& \multirow{2}{*}{Ray-finned Fishes} & $acc$ & $45.4$ & \multicolumn{1}{c|}{$50.8$} & $45.7$ & \multicolumn{1}{c|}{$53.8$} & $39.4$ & \multicolumn{1}{c|}{$47.8$} & $37.7$ & $44.1$ \\
&& $f$ & $0.07$ & \multicolumn{1}{c|}{$0.06$} & $0.07$ & \multicolumn{1}{c|}{$0.07$} & $0.07$ & \multicolumn{1}{c|}{$0.07$} & $0.08$ & $0.08$ \\ \cline{3-11}
& \multirow{2}{*}{Amphibians} & $acc$ & $42.4$ & \multicolumn{1}{c|}{$51.7$} & $48.5$ & \multicolumn{1}{c|}{$52$} & $41.9$ & \multicolumn{1}{c|}{$48.5$} & $34.7$ & $39$ \\
&& \cellcolor{red!40}$f$ & \cellcolor{red!40}$0.14$ & \multicolumn{1}{c|}{\cellcolor{red!40}$0.14$} & \cellcolor{red!40}$0.15$ & \multicolumn{1}{c|}{\cellcolor{red!40}$0.14$} & \cellcolor{red!40}$0.14$ & \multicolumn{1}{c|}{\cellcolor{red!40}$0.15$} & \cellcolor{red!40}$0.14$ & \cellcolor{red!40}$0.15$ \\ \cline{3-11}
& \multirow{2}{*}{Mollusks} & $acc$ & $43.2$ & \multicolumn{1}{c|}{$52.8$} & $43.2$ & \multicolumn{1}{c|}{$55.3$} & $36.2$ & \multicolumn{1}{c|}{$49.4$} & $33.9$ & $44.7$ \\
&& $f$ & $0.13$ & \multicolumn{1}{c|}{$0.13$} & $0.13$ & \multicolumn{1}{c|}{$0.14$} & $0.12$ & \multicolumn{1}{c|}{$0.13$} & $0.12$ & $0.13$ \\ \cline{3-11}
& \multirow{2}{*}{Arachnids} & $acc$ & $42.9$ & \multicolumn{1}{c|}{$50.6$} & $42.8$ & \multicolumn{1}{c|}{$53.1$} & $38.1$ & \multicolumn{1}{c|}{$47.6$} & $29$ & $41.2$ \\
&& $f$ & $0.07$ & \multicolumn{1}{c|}{$0.06$} & $0.06$ & \multicolumn{1}{c|}{$0.07$} & $0.07$ & \multicolumn{1}{c|}{$0.06$} & $0.06$ & $0.07$ \\ \cline{3-11}
& \multirow{2}{*}{Animalia} & $acc$ & $44.3$ & \multicolumn{1}{c|}{$51.5$} & $41.8$ & \multicolumn{1}{c|}{$55.5$} & $40.2$ & \multicolumn{1}{c|}{$45.8$} & $30.7$ & $39.1$ \\
&& \cellcolor{red!40}$f$ & \cellcolor{red!40}$0.15$ & \multicolumn{1}{c|}{\cellcolor{red!40}$0.17$} & \cellcolor{red!40}$0.15$ & \multicolumn{1}{c|}{\cellcolor{red!40}$0.17$} & \cellcolor{red!40}$0.15$ & \multicolumn{1}{c|}{\cellcolor{red!40}$0.16$} & \cellcolor{red!40}$0.14$ & \cellcolor{red!40}$0.15$ \\ \hline\hline 
\multicolumn{2}{l}{All-class average} & $\mu_{\mathrm{acc}}$ & $43.4$ & \multicolumn{1}{c|}{$51.2$} & $44.4$ & \multicolumn{1}{c|}{$53.5$} & $38.3$ & \multicolumn{1}{c|}{$47$} & $34.1$ & \multicolumn{1}{c}{$40.5$}\\
\multicolumn{2}{l}{All-class variance} & $\sigma^2_{\mathrm{acc}}$ & $3.5$ & \multicolumn{1}{c|}{$1.9$} & $5.5$ & \multicolumn{1}{c|}{$3.7$} & $3.5$ & \multicolumn{1}{c|}{$4.5$} & $7.4$ & \multicolumn{1}{c}{$4.7$}
\end{tabular}}
	\label{tab:find_ratios_inat21}
\end{table*}

In this publication, BiCDO is explained and applied using accuracy. However, it can also be used with any other objective functions, depending on which ones need to be optimized. All objective functions that consider only individual classes, e.g. accuracy, precision and recall are possible. Objective functions that also take into account correlations between classes, or that cannot be summarised in a single value, will not give meaningful results. This includes, for example, the F1 score and the precision-recall curve.

\section{Experiments}
\label{sec:experiments}
This chapter presents the experiments and their results. In \cref{subsec:find_ratios} we show in detail why this method works independently of the architecture of the multi-class model and the multi-class dataset, as long as there is one label per image. Furthermore, in \cref{subsec:expand_ratios} we will show that the found class-based distributions can be expanded relative to each other to optimize the accuracy of the model. We show this using the EfficientNet of sizes b0 and b3 developed by \cite{Tan.2019}, the ResNet from \cite{He.2016} and the ConvNeXt from \cite{Liu.2022}. As multi-class datasets we use the CIFAR-10 dataset (\cite{A.Krizhevsky.2009}) and the iNaturalist21 dataset (\cite{inat.2021}). All following results were performed on the validation dataset. All reductions and increases in the amount of data refer to the training dataset. See the supplementary material for details on preprocessing steps and other important parameters.

\begin{table*}[tbh]
	\caption{The table shows the rounded CIFAR-10 results for EfficientNet-B3 model with fixed parameters: $\boldsymbol{f_c=0.5}$, $\boldsymbol{l_c=0.05}$, $\boldsymbol{u_c=0.95}$, $\boldsymbol{D_c=2000}$, over 150 iterations. Target offsets $o_c$ are applied to certain classes. Highlighted cells show class accuracies ($\boldsymbol{\%}$) affected by these offsets.}
    \centering
    \small
    \makebox[\linewidth]{
\begin{tabular}{l|ll||cccccccc}
\multicolumn{3}{c}{} & \multicolumn{2}{c}{(A)} & \multicolumn{2}{c}{(B)} & \multicolumn{2}{c}{(C)} & \multicolumn{2}{c}{(D)} \\
\multicolumn{3}{c||}{\multirow{3}{*}{Target offset $o_c$}} & \multicolumn{1}{l}{$o_{\mathrm{Automobile}}$} & \multicolumn{1}{l|}{$=0.1$} & \multicolumn{1}{l}{$o_{\mathrm{Frog}}$} & \multicolumn{1}{l|}{$=-0.05$} & \multicolumn{1}{l}{$o_{\mathrm{Airplane}}$} & \multicolumn{1}{l|}{$=-0.05$} & \multicolumn{1}{l}{$o_{\mathrm{Deer}}$} & \multicolumn{1}{l}{$=0.1$} \\
\multicolumn{3}{c||}{} &  & \multicolumn{1}{c|}{} &  & \multicolumn{1}{c|}{} & \multicolumn{1}{l}{$o_{\mathrm{Cat}}$} & \multicolumn{1}{l|}{$=0.05$} & \multicolumn{1}{l}{$o_{\mathrm{Horse}}$} & \multicolumn{1}{l}{$=-0.15$} \\
\multicolumn{3}{c||}{} & \multicolumn{1}{l}{$o_{\mathrm{else}}$} & \multicolumn{1}{l|}{$=0$} & \multicolumn{1}{l}{$o_{\mathrm{else}}$} & \multicolumn{1}{l|}{$=0$} & \multicolumn{1}{l}{$o_{\mathrm{else}}$} & \multicolumn{1}{l|}{$=0$} & \multicolumn{1}{l}{$o_{\mathrm{else}}$} & \multicolumn{1}{l}{$=0$} \\ \hline\hline
\multirow{20}{*}{\rotatebox[origin=c]{90}{Class results}} & \multirow{2}{*}{Airplane} & $acc$ &  \multicolumn{2}{c|}{$59.2$} & \multicolumn{2}{c|}{$60.2$} & \multicolumn{2}{c|}{\cellcolor{red!40}$56.2$} & \multicolumn{2}{c}{$60.2$} \\
&& $f$ & \multicolumn{2}{c|}{$0.05$} & \multicolumn{2}{c|}{$0.06$} & \multicolumn{2}{c|}{$0.05$} & \multicolumn{2}{c}{$0.05$} \\ \cline{3-11}
& \multirow{2}{*}{Automobile} & $acc$ & \multicolumn{2}{c|}{\cellcolor{red!40}$68.2$} & \multicolumn{2}{c|}{$62.2$} & \multicolumn{2}{c|}{$58.2$} & \multicolumn{2}{c}{$58.9$} \\
&& $f$ & \multicolumn{2}{c|}{$0.05$} & \multicolumn{2}{c|}{$0.03$} & \multicolumn{2}{c|}{$0.03$} & \multicolumn{2}{c}{$0.03$} \\ \cline{3-11}
& \multirow{2}{*}{Bird} & $acc$ &  \multicolumn{2}{c|}{$61.7$} & \multicolumn{2}{c|}{$61.1$} & \multicolumn{2}{c|}{$55.6$} & \multicolumn{2}{c}{$60.2$} \\
&& $f$ & \multicolumn{2}{c|}{$0.15$} & \multicolumn{2}{c|}{$0.15$} & \multicolumn{2}{c|}{$0.15$} & \multicolumn{2}{c}{$0.16$} \\ \cline{3-11}
& \multirow{2}{*}{Cat} & $acc$ &  \multicolumn{2}{c|}{$59.7$} & \multicolumn{2}{c|}{$63$} & \multicolumn{2}{c|}{\cellcolor{red!40}$59.8$} & \multicolumn{2}{c}{$61.7$} \\
&& $f$ & \multicolumn{2}{c|}{$0.23$} & \multicolumn{2}{c|}{$0.24$} & \multicolumn{2}{c|}{$0.27$} & \multicolumn{2}{c}{$0.23$} \\ \cline{3-11}
& \multirow{2}{*}{Deer} & $acc$ &  \multicolumn{2}{c|}{$60.3$} & \multicolumn{2}{c|}{$62.3$} & \multicolumn{2}{c|}{$55.9$} & \multicolumn{2}{c}{\cellcolor{red!40}$71.7$} \\
&& $f$ & \multicolumn{2}{c|}{$0.11$} & \multicolumn{2}{c|}{$0.12$} & \multicolumn{2}{c|}{$0.11$} & \multicolumn{2}{c}{$0.17$} \\ \cline{3-11}
& \multirow{2}{*}{Dog} & $acc$ &  \multicolumn{2}{c|}{$59.9$} & \multicolumn{2}{c|}{$61$} & \multicolumn{2}{c|}{$56.9$} & \multicolumn{2}{c}{$58.8$} \\
&& $f$ & \multicolumn{2}{c|}{$0.2$} & \multicolumn{2}{c|}{$0.2$} & \multicolumn{2}{c|}{$0.19$} & \multicolumn{2}{c}{$0.18$} \\ \cline{3-11}
& \multirow{2}{*}{Frog} & $acc$ &  \multicolumn{2}{c|}{$59.8$} & \multicolumn{2}{c|}{\cellcolor{red!40}$57.7$} & \multicolumn{2}{c|}{$56.7$} & \multicolumn{2}{c}{$58.5$} \\
&& $f$ & \multicolumn{2}{c|}{$0.07$} & \multicolumn{2}{c|}{$0.06$} & \multicolumn{2}{c|}{$0.07$} & \multicolumn{2}{c}{$0.07$} \\ \cline{3-11}
& \multirow{2}{*}{Horse} & $acc$ &  \multicolumn{2}{c|}{$60$} & \multicolumn{2}{c|}{$60.4$} & \multicolumn{2}{c|}{$56.4$} & \multicolumn{2}{c}{\cellcolor{red!40}$42.8$} \\
&& $f$ & \multicolumn{2}{c|}{$0.07$} & \multicolumn{2}{c|}{$0.07$} & \multicolumn{2}{c|}{$0.07$} & \multicolumn{2}{c}{$0.03$} \\ \cline{3-11}
& \multirow{2}{*}{Ship} & $acc$ &  \multicolumn{2}{c|}{$57.8$} & \multicolumn{2}{c|}{$66.2$} & \multicolumn{2}{c|}{$54.6$} & \multicolumn{2}{c}{$59$} \\
&& $f$ & \multicolumn{2}{c|}{$0.04$} & \multicolumn{2}{c|}{$0.04$} & \multicolumn{2}{c|}{$0.03$} & \multicolumn{2}{c}{$0.04$} \\ \cline{3-11}
& \multirow{2}{*}{Truck} & $acc$ &  \multicolumn{2}{c|}{$60.8$} & \multicolumn{2}{c|}{$60.8$} & \multicolumn{2}{c|}{$56.7$} & \multicolumn{2}{c}{$64.6$} \\
&& $f$ & \multicolumn{2}{c|}{$0.03$} & \multicolumn{2}{c|}{$0.03$} & \multicolumn{2}{c|}{$0.03$} & \multicolumn{2}{c}{$0.03$} \\ \hline\hline 
\multicolumn{2}{l}{All-class average} & $\mu_{\mathrm{acc}}$ &  \multicolumn{2}{c|}{$60.7$} & \multicolumn{2}{c|}{$61.5$} & \multicolumn{2}{c|}{$56.7$} & \multicolumn{2}{c}{$59.6$}\\
\multicolumn{2}{l}{All-class variance} & $\sigma^2_{\mathrm{acc}}$ &  \multicolumn{2}{c|}{$7.1$} & \multicolumn{2}{c|}{$4.4$} & \multicolumn{2}{c|}{$1.9$} & \multicolumn{2}{c}{$45.9$}
\end{tabular}}
	\label{tab:find_ratios_target_offset}
\end{table*}
\renewcommand{\arraystretch}{1.1}
\begin{table*}[tbh]
	\caption{This table shows the final results, which are based on distribution factors ($\boldsymbol{f_c}$), the maximum available data. Highlights in the first two columns indicate whether the BiCDO approach or the traditional approach performs better. The final column indicates where accuracies ($\boldsymbol{\%}$) should exceedg the mean by 10\%.}
    \centering
    \small
    \makebox[\linewidth]{
\begin{tabular}{l|l||cccccc}
\multicolumn{2}{c}{} & \multicolumn{2}{c}{(A)} & \multicolumn{2}{c}{(B)} & \multicolumn{2}{c}{(C)} \\
\multicolumn{2}{c||}{\multirow{2}{*}{Used $f_c$}} & \multicolumn{2}{c|}{Equally distributed} & \multicolumn{2}{c|}{(B).2k of \cref{tab:find_ratios_cifar10}} & \multicolumn{2}{c}{(A) of \cref{tab:find_ratios_target_offset}} \\ 
\multicolumn{2}{c||}{} & \multicolumn{1}{c}{$acc$} & \multicolumn{1}{c|}{$num$} & \multicolumn{1}{c}{$acc$} & \multicolumn{1}{c|}{$num$} & \multicolumn{1}{c}{$acc$} & \multicolumn{1}{c}{$num$} \\ \hline\hline
\multirow{10}{*}{\rotatebox[origin=c]{90}{Class results}} & \multirow{1}{*}{Airplane} & \multicolumn{1}{c}{\cellcolor{red!40}$80.3$} & \multicolumn{1}{c|}{$1980$} & \multicolumn{1}{c}{$78.6$} & \multicolumn{1}{c|}{$1005$} & \multicolumn{1}{c}{$78.2$} & \multicolumn{1}{c}{$1089$} \\
& \multirow{1}{*}{Automobile} & \multicolumn{1}{c}{\cellcolor{red!40}$86.7$} & \multicolumn{1}{c|}{$2009$} & \multicolumn{1}{c}{$80.6$} & \multicolumn{1}{c}{$506$} & \multicolumn{1}{c}{\cellcolor{red!40}$88.3$} & \multicolumn{1}{c}{$1023$} \\
& \multirow{1}{*}{Bird} & \multicolumn{1}{c}{$68.4$} & \multicolumn{1}{c|}{$1942$} & \multicolumn{1}{c}{\cellcolor{red!40}$78$} & \multicolumn{1}{c|}{$2936$} & \multicolumn{1}{c}{$79.2$} & \multicolumn{1}{c}{$3294$} \\
& \multirow{1}{*}{Cat} & \multicolumn{1}{c}{$56.6$} & \multicolumn{1}{c|}{$1990$} & \multicolumn{1}{c}{\cellcolor{red!40}$77.2$} & \multicolumn{1}{c|}{$5000$} & \multicolumn{1}{c}{$76.9$} & \multicolumn{1}{c}{$5000$} \\
& \multirow{1}{*}{Deer} & \multicolumn{1}{c}{$72.9$} & \multicolumn{1}{c|}{$1865$} & \multicolumn{1}{c}{\cellcolor{red!40}$76.9$} & \multicolumn{1}{c|}{$2216$} & \multicolumn{1}{c}{$79.1$} & \multicolumn{1}{c}{$2476$} \\
& \multirow{1}{*}{Dog} & \multicolumn{1}{c}{$67.1$} & \multicolumn{1}{c|}{$2027$} & \multicolumn{1}{c}{\cellcolor{red!40}$75.2$} & \multicolumn{1}{c|}{$4039$} & \multicolumn{1}{c}{$74.8$} & \multicolumn{1}{c}{$4234$} \\
& \multirow{1}{*}{Frog} & \multicolumn{1}{c}{\cellcolor{red!40}$82.9$} & \multicolumn{1}{c|}{$2053$} & \multicolumn{1}{c}{$78.5$} & \multicolumn{1}{c|}{$1248$} & \multicolumn{1}{c}{$80.8$} & \multicolumn{1}{c}{$1547$} \\
& \multirow{1}{*}{Horse} & \multicolumn{1}{c}{\cellcolor{red!40}$81.1$} & \multicolumn{1}{c|}{$2005$} & \multicolumn{1}{c}{$79$} & \multicolumn{1}{c|}{$1596$} & \multicolumn{1}{c}{$77.9$} & \multicolumn{1}{c}{$1436$} \\
& \multirow{1}{*}{Ship} & \multicolumn{1}{c}{\cellcolor{red!40}$85.8$} & \multicolumn{1}{c|}{$1933$} & \multicolumn{1}{c}{$80.2$} & \multicolumn{1}{c|}{$656$} & \multicolumn{1}{c}{$82.2$} & \multicolumn{1}{c}{$800$} \\
& \multirow{1}{*}{Truck} & \multicolumn{1}{c}{\cellcolor{red!40}$85.8$} & \multicolumn{1}{c|}{$1933$} & \multicolumn{1}{c}{$79.2$} & \multicolumn{1}{c|}{$538$} & \multicolumn{1}{c}{$78.1$} & \multicolumn{1}{c}{$728$} \\ \hline\hline
\multicolumn{2}{l||}{$\mu_{\mathrm{acc}}\pm\sigma^2_{\mathrm{acc}}$ and $\sum_{\mathrm{num}}$} &  \multicolumn{1}{c}{$76.6\pm88.7$} & \multicolumn{1}{c|}{$19737$} & \multicolumn{1}{c}{\cellcolor{red!40}$78.3\pm2.3$} & \multicolumn{1}{c|}{$19740$} & \multicolumn{1}{c}{$79.5\pm12.6$} & \multicolumn{1}{c}{$21627$} \\
\end{tabular}}
	\label{tab:train_ratios}
\end{table*}
To find the Pareto optimized class-based data distributions, the BiCDO method is applied to an existing training pipeline. The framework is started with different numbers of maximum images to analyse how many images are required for a meaningful use of BiCDO. \Cref{fig:experiments_find} shows the BiCDO process for finding the distribution factor of each class for the result shown in \cref{fig:introduction}, which is a comparison between an equal and an BiCDO distribution (for the numerical results see the highlighted column (B).2k in \cref{tab:find_ratios_cifar10}). For this run, the training pipeline uses the EfficientNet of size b3 and the CIFAR-10 dataset. This run was performed for all classes with the start parameters $D_c=2000$, $f_c=0.5$, $l_c=0.05$, $u_c=0.95$ and $o_c=0$. It can be seen that the distribution adapts with each trained model until it converges after about $100$ iterations. A similar convergence behaviour can also be observed when looking at the corresponding accuracy in \cref{fig:experiments_find}. Further experiments were performed, varying the number of $D_c$ on different model architectures (for each class $c$ to the same value). The results after $150$ iterations are shown in \cref{tab:find_ratios_cifar10}.

It can be seen that no matter what model architecture, model size or how much data $D_c$ ($=500$, $=2000$, $=5000$) per class is provided to BiCDO, very similar class ratios $f_c$ are always found for each class. In addition, the accuracy of each class converges over the training runs, so that the maximum variance is $4.4$ in column (A).0.5k in \cref{tab:find_ratios_cifar10}. Initially, all classes were given the same amount of data $D_c$. However, the use of BiCDO has shown that some classes need considerably more images to achieve comparable accuracy. For example, the Cat class with a distribution factor of $f_{\mathrm{Cat}}\geq0.2$ always requires at least $5$ times as much data as the Automobile class with a distribution factor of $f_{\mathrm{Automobile}}\leq0.04$ (see the highlighted rows in \cref{tab:find_ratios_cifar10}). This allows us to determine fairly accurately how much data per class is required to achieve a given (higher) level of accuracy. This is taken up again in \cref{subsec:expand_ratios} and demonstrated by tests. As a result, uniform accuracy across all classes depends very much on the actual data and not so much on the architecture of the model. The architecture, on the other hand, is responsible for making more or less optimal use of the data.

\Cref{tab:find_ratios_inat21} shows further tests on the iNaturalist21 dataset with the same starting parameters. It shows that the approach works well for supposedly harder problems. Even with the iNaturalist21 dataset as a base, very similar accuracy is achieved for each class. The distribution factors $f_c$ are also built in such a way that many classes require different numbers of images to achieve this accuracy. A closer look at the results of the distribution factors $f_c$ shows that the classes Reptiles, Amphibians and Animalia always require at least $2.3$ as many images as the classes Plants, Birds and Fungi (see the highlighted rows in \cref{tab:find_ratios_inat21}).

So far we have mainly argued that all classes should have a similar accuracy target. However, this may not always be the case when it comes to safety-critical functions, or when different classes have different priorities. The target offset $o_c$ has been introduced to allow BiCDO to be used in this case. The \cref{tab:find_ratios_target_offset} shows the results for the EfficientNet of size b3 and the fixed start parameters $f_c=0.5$, $l_c=0.05$, $u_c=0.95$ and $D_c=2000$ for the CIFAR-10 dataset and $150$ iterations. The parameter $o_c$ has been set to different values to show the functionality.
It can be seen how the target offset $o_c$ is taken into account by the factors $f_c$ so that it corresponds to the previously defined requirements. The highlighted cells in \cref{tab:find_ratios_target_offset} show how the accuracy has actually aligned for this class and how it differs from the all-class average in the expected direction compared to the other classes. Compared to the run with the same parameters and the target offset $o_c=0$ (column (B).2k of \cref{tab:find_ratios_cifar10}), it can be seen that the all-class average has decreased by about $10$ points. This shows, that the distribution found with $o_c=0$ is Pareto optimized. However, this is no longer relevant as the corresponding distribution factors $f_c$ have been calculated for this accuracy and can be used for further work. This behaviour can be confirmed for all the tests performed. 

\subsection{Expand Pareto Optimized Distribution}
\label{subsec:expand_ratios}
The Pareto optimized class-based distribution factors can be found using BiCDO. This is independent of whether a particular class should have a target offset $o_c$ or not. In this chapter we show that the distribution factors $f_c$ can be used to improve the accuracy for each class uniformly. This can be achieved by increasing the amount of data for each class with respect to the distribution factors $f_c$. In addition, we will show that BiCDO actually generates Pareto optimized class-based data distributions by dividing the same total amount of data equally across all classes.  The distribution factors $f_c$ in \cref{tab:find_ratios_cifar10} to \cref{tab:find_ratios_target_offset} show how much data each class requires relative to the others. For example, in the column (B).2k in \cref{tab:find_ratios_cifar10} you can see that the class Cat requires the most data. In this new experiment, all images available for this class in the dataset ($5000$) are now made available. All other classes are adjusted relative to it, so as not to change the Pareto optimized factors. The Car class is therefore given $(f_{\mathrm{Automobile}}/f_{\mathrm{Cat}})\cdot5000=506$ images. This calculation is based on the unrounded values. A further training is done using the same principle, based on the distribution factors $f_c$ in \cref{tab:find_ratios_target_offset} in column (A). There is also a training run where the total amount of data is uniformly distributed across all classes. The results of the three training runs, including the number of images used for each class ($num$), are shown in \cref{tab:train_ratios}.

The first two columns (A) and (B) of \cref{tab:train_ratios} are the two test runs that can also be seen in \cref{fig:introduction} in \cref{sec:introduction}. This shows that a uniform distribution of a dataset as in the column (A) leads to slightly worse overall accuracy than the BiCDO optimized distribution in the column (B). This is already an improvement. If the minimum and maximum values are also considered, it becomes clear that the class-specific accuracy is much more uniform with the BiCDO optimized distribution. Even the variance with $3$ of the accuracy in finding the distribution factors $f_c$ (column (B).2k of \cref{tab:find_ratios_cifar10}) is decreasing to $2.3$ when the distribution factors $f_c$ are applied to all available images (column (B) of \cref{tab:train_ratios}). When finding the Pareto optimized class-based data distribution in column (B).2k of \cref{tab:find_ratios_cifar10}, $7488$ images were used in the end. In the extended experiment in \cref{tab:train_ratios} column (B), $19740$ images were used to improve the performance. The relative distances between the accuracies of the classes remained almost the same. This suggests that the approach can be run with a small amount of data and then applied to much more data. This is also true for the approach with the target offset $o_c$. Looking at column (C) in \cref{tab:train_ratios}, it can be seen that the car class is still about 10\% above the average. 

\section{Discussion}
\label{sec:discussion}
The developed BiCDO framework offers several advantages. It can be used to determine the Pareto optimal distributions between the classes of the dataset in order to obtain a balanced distribution between the classes in a defined objective function. In our experiments, the optimized distribution was always a systematic long-tail distribution (compare each column of \cref{tab:find_ratios_cifar10,tab:find_ratios_inat21,tab:find_ratios_target_offset}). This shows that a uniform distribution between the classes of a dataset does not necessarily lead to a uniform value of the classes in the objective function. A prerequisite for its use is the ready definition of the classes to be detected and a basic set of data for each class. An already defined architecture is preferred, but does not necessarily have to be available, as we have experimented. Once the requirements are met, BiCDO can be used for data collection. It is possible to predict how many images will be needed to increase the accuracy of all classes equally. Ultimately, this can save a lot of money and time when developing models that require a lot of data, and help to optimize a dataset. Furthermore, BiCDO can be used to demonstrate to current norms or standards, such as the EU AI Act, that a sufficient amount of data has been used for the intended purpose for each class, as there is a small variance between the values of the classes in the objective function.

In addition to the benefits, there is one hurdle that needs to be overcome or accepted for BiCDO to be used successfully. This is the runtime. In our tests, we always scheduled 150 iterations for BiCDO. This means that a model has to be trained 150 times. This can take several days, if not weeks. In our experience, this effort in time, labour and CO2 is outweighed by the savings and other benefits of using BiCDO. However, it should still be considered whether the framework is worthwhile in the specific application. The effort can be adjusted to some extent by not training the model to the end, as BiCDO follows an iterative approach, only a tendency is needed to take an optimization step in the right direction. In addition, it would be possible to add a convergence factor to BiCDO to make the framework converge with larger steps at the beginning.

Another point to consider is that image classification covers only a small part of the applications of models in computer vision. In some other areas, such as object detection and semantic segmentation, BiCDO cannot yet be used in this form. BiCDO currently exploits the fact that each image can be assigned to exactly one class. This allows the exact amount of data to be determined. As soon as objects are to be considered, it is possible that many objects occur in an image. This means that a new optimization problem has to be solved when providing the data. In this solution, it should be noted that there should be a random proportion in the selection of images in order to follow the framework and not always provide the same images. In addition, a non-computationally intensive optimization algorithm should be applied based on this. In this way, the problem of providing the data can be solved as well as possible without significantly increasing the time problem mentioned above.

\textbf{Conclusion}. In this paper we have introduced the new BiCDO framework. We have shown that BiCDO can be used to align the objective function for each class to a specific, but undefined, value. This results in a Pareto optimized distribution between all classes. We have also shown that this framework can be used architecture and dataset independent. To cover demands from industrial and safety-critical functions, we introduced a target offset $o_c$, which can be used to give individual classes a relative increase in accuracy compared to the average accuracy of all classes. Contrary to conventional wisdom, our approach shows that an uniform distribution of classes in the dataset often has a negative effect on the final result and the trained models are often subject to an unintended bias. In all of our tests, BiCDO achieved a long-tail distribution as the optimal amount of data. This suggests that a systematic long-tail distribution is the optimal distribution for many problems.

\printbibliography

\end{document}